\documentclass[times,twocolumn,final,authoryear]{elsarticle}

\usepackage{prletters}
\usepackage{framed,multirow}

\usepackage{amsmath}
\usepackage{mathrsfs}
\usepackage{graphicx}
\usepackage{subcaption}
\usepackage{bm}

\usepackage{booktabs}       

\usepackage{url}
\usepackage{xcolor}
\definecolor{newcolor}{rgb}{.8,.349,.1}

\journal{Pattern Recognition Letters}

\begin{document}

\thispagestyle{empty}

\clearpage
\thispagestyle{empty}
\ifpreprint
  \vspace*{-1pc}
\fi

\begin{frontmatter}

\title{Encoding the Local Connectivity Patterns of fMRI for Cognitive State Classification }

\author[1]{Itir \snm{Onal Ertugrul}\corref{cor1}} 
\cortext[cor1]{Corresponding author: 
  Tel.: +90-555-557-9677;  
}
\ead{itir@ceng.metu.edu.tr}
\author[2]{Mete \snm{Ozay}}
\author[1]{Fatos T. \snm{Yarman Vural}}

\address[1]{Department of Computer Engineering, Middle East Technical University, Ankara, Turkey}
\address[2]{Graduate School of Information Sciences, Tohoku University, Sendai, Miyagi, Japan}

\received{1 May 2013}
\finalform{10 May 2013}
\accepted{13 May 2013}
\availableonline{15 May 2013}
\communicated{S. Sarkar}

\begin{abstract}

In this work, we propose a novel framework to encode the local connectivity patterns of brain, using Fisher Vectors (FV), Vector of Locally Aggregated  Descriptors (VLAD) and Bag-of-Words (BoW) methods. We first obtain local descriptors, called Mesh Arc Descriptors (MADs) from fMRI data, by forming local meshes around anatomical regions, and estimating their relationship within a neighborhood. Then, we extract a dictionary of relationships, called \textit{brain connectivity dictionary} by fitting a generative Gaussian mixture model (GMM) to a set of MADs, and selecting the codewords at the mean of each component of the mixture. Codewords represent the connectivity patterns among anatomical regions. We also encode MADs by VLAD and BoW methods using the k-Means clustering.

We classify the cognitive states of Human Connectome Project (HCP) task fMRI dataset, where we train support vector machines (SVM) by the encoded MADs. Results demonstrate that, FV encoding of MADs can be successfully employed for classification of cognitive tasks, and outperform the  VLAD and BoW representations. Moreover, we identify the significant Gaussians in mixture models by computing energy of their corresponding FV parts, and analyze their effect on classification accuracy. Finally, we suggest a new method to visualize the codewords of brain connectivity dictionary.

\end{abstract}


\end{frontmatter}


\section{Introduction}
\label{sec1}
Functional Magnetic Resonance Imaging (fMRI) is a powerful tool used for capturing the neural activation observed in a wide range of cognitive tasks encoded in brain. Traditional approaches employ fMRI Blood Oxygenation Level Dependent (BOLD) response of voxels or anatomical regions for cognitive state classification \citep{Haxby2001,mitchell2004}. Yet, recent findings show that connectivity between voxels or anatomical regions provides more information about activities in brain compared to the voxel BOLD responses. In addition, pairwise correlations between voxel BOLD responses are suggested to represent and classify cognitive tasks \citep{Richiardi2011,Shirer2011}. Pairwise global connectivity provides better classification performance compared to traditional approaches. Brain connectivity is also represented by a set of local meshes \citep{Onal2015_1,Onal2015_2}, where relationships among multiple voxels are estimated within a predefined neighborhood. Estimated relationships, called Mesh Arc Descriptors \citep{mads}, are reported to give the best performance in cognitive state classification, compared to voxel BOLD responses and their pairwise relationships.


Encoding approaches are widely used in pattern recognition literature to improve the representation power of local descriptors. In a popular encoding approach, called Bag-of-Words (BoW), first local descriptors are clustered using the k-means clustering method. Then, cluster centroids are defined as codewords to form a dictionary.  Codewords are utilized as \textit{textual words} in natural language processing, and \textit{visual words} in image processing. In this approach, each descriptor is assigned to the closest codeword, and a sample is represented by a histogram of codewords, to which its local descriptors belong.   BoW approach has been used to detect diseases using fMRI data \citep{Solmaz2012}, and to classify EEG time series data \citep{Wang2013,Merino2013}. 

In another encoding approach, called Vector of Locally Aggregated Descriptors (VLAD), first, codewords are computed similar to BoW. However, VLAD aims to accumulate the difference between the codewords and local descriptors assigned to the codewords \citep{Jegou2010}. VLAD has been used in various applications including image \citep{Delhumeau2013} and video \citep{Abbas2015} processing, yet it has not been used to encode local descriptors obtained from fMRI data.

Fisher Vector (FV) encoding methods are employed for statistical data analysis by making use of the benefits of generative and discriminative models \citep{Carvajal2016,Sanchez2013}.  FV encoding is considered as an extension of BoW such that, rather than encoding the relative  frequency of the descriptors, it encodes the information on  distribution of the descriptors \citep{Perronnin2010}. 
Fisher kernels have been used to compute FVs utilizing a mechanism that incorporates generative probability models into discriminative classifiers in applications, including classification of protein domains \citep{Jaakkola1999}, action and event recognition \citep{Oneta2013,Sekma2015}, image classification \citep{Liu2014,Simonyan2013,sanchez2015} and 3D object retrieval \citep{Savelonas2016}.
FVs have been applied to model effective connectivity of the networks using MRI and PET data \citep{Zhou2016}. Yet, FVs have not been used to encode the connectivity patterns of fMRI data, which is very crucial to analyze the behavior of brain during cognitive tasks.

The major contribution of this study is to suggest a novel framework for encoding a set of local descriptors which models the connectivity patterns among the anatomic regions, based on the fMRI data. Inspired by the concept of visual words in image processing, our motivation is to generate a \textit{brain connectivity dictionary} that represents relational patterns formed among anatomical regions in order to describe the cognitive states. The proposed framework enables us to examine various encoding methods such as FVs, VLAD and BoW for cognitive task classification. To our knowledge, our framework is the first that uses encoding methods to the connectivity analysis of fMRI data. 


In order to generate the \textit{brain connectivity dictionary}, we fit a Gaussian mixture model (GMM) to the set of MADs, obtained from all anatomic regions, and from all subjects, during all cognitive tasks. The mean vector of a Gaussian component corresponds to a codeword of \textit{brain connectivity dictionary}. Then, by using the codewords, we encode MADs with FV, VLAD and BoW methods. In FV encoding, a generative model is estimated by fitting a Gaussian mixture to the MAD vectors. We represent each sample of a cognitive task, as a FV encoding to model the statistical deviation of the associated MAD from the corresponding GMMs. In VLAD, we cluster MADs using a k-means clustering method and obtain a dictionary of MADs. We represent each sample by the accumulated difference between the MADs and the codewords to which they are assigned. For comparison, we also compute BoW models of MAD. We train linear SVM classifiers by the encoded MADs, obtained from BoW, VLAD and FV encoding methods. We observe that encoding MADs with FV provides the best performances and improves the performances of raw MADs in cognitive task classification. On the other hand, VLAD encoding provides similar performance to that of raw MADs. However, BoW encoding decreases the performance of raw MADs. Our results show that the performance of SVM classifier heavily depends on the modality of the codewords, the dimension of feature vectors and degree of functional locality of meshes.

Another contribution of this work is the exploration of the relationship between energy of FV columns obtained from a particular Gaussian component  and the classification accuracy. Our results reflect that Gaussian mixture components whose FV columns have higher energy are more discriminative compared to the ones with lower energy.  Finally, we suggest a visualization method to depict and analyze codewords of the brain connectivity dictionary on a human brain template. We first sort the Gaussian mixture components based on the energy they provide for FV encoding. Then, we plot their codewords on brain, and sort them with respect to their discriminative power. The suggested framework is tested to classify the cognitive tasks in Human Connectomme Project (HCP). The results suggest that MADs having large values in the elements corresponding to occipital regions and low values in the rest of the elements, provide the most significant information to classify the cognitive tasks. On the other hand, MADs having large values in the elements corresponding to central structures (Putamen, Caudate and Thalamus) provide the least discriminative information.

\begin{table}[h!]
	\centering
	\caption{Number of scans obtained per session, and its duration (min:sec).}
	\label{tab:7tasks}   
	\begin{tabular}{lccccccc}
		\hline
		& \textbf{Emo}& \textbf{Gam} & \textbf{Lan} & \textbf{Mot} & \textbf{Rel} & \textbf{Soc} & \textbf{WM} \\ \hline
		\textbf{Scans}    & 176              & 253               & 316               & 284            & 232                 & 274             & 405         \\ 
		\textbf{Dur}. & 2:16             & 3:12              & 3:57              & 3:34           & 2:56                & 3:27            & 5:01        \\ \hline
	\end{tabular}
    \vspace{-0.3cm}
\end{table}

\begin{figure*}[t]
	\centering
	\includegraphics[scale=0.34]{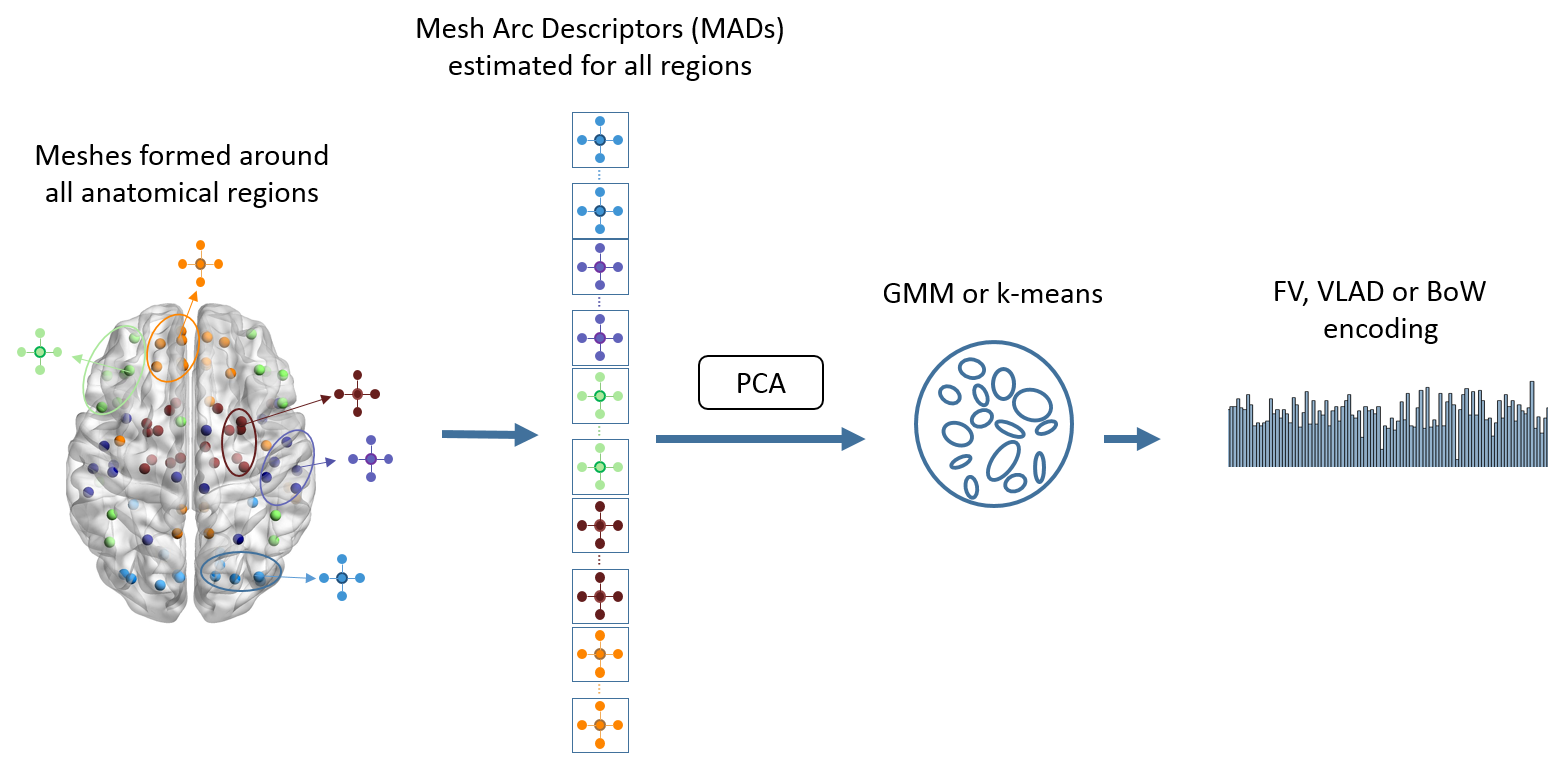}
	\caption{An overview of our proposed framework which is employed to generate a brain connectivity dictionary. For each cognitive task, we first form local meshes around all anatomical regions using their functionally nearest neighbors (in the figure, we represent only five representative meshes). 
		Next, we estimate  edge weights of meshes (MAD) for all tasks of all subjects. We apply PCA to reduce dimension of the MADs, and obtain uncorrelated descriptors. Then, we employ GMMs and k-means clustering methods on the descriptors using data obtained from all tasks. Finally, we obtain FV, VLAD and BoW encoding for each task.}
	\label{fig:flowchart}
\end{figure*}

\section{Data Representation}

We use the benchmark task fMRI dataset of Human Connectome Project (HCP) \citep{Barch2013}. We employ task fMRI data of 97 healthy subjects collected for seven cognitive tasks, namely, Emotion (Emo), Gambling (Gam), Language Processing (Lan), Motor (Mot), Relational Processing (Rel), Social Processing (Soc) and Working Memory (WM). Number of scans and their duration vary for each task, yet they have equal duration for all participants (see Table \ref{tab:7tasks}).

We use $R = 98$ regions of Automated Anatomical Labeling (AAL) brain atlas \cite{Tzourio2002} except the ones that cover Cerebellum. Each region is denoted by $r_i$, where $i = 1, 2, \ldots ,R$. Note that, each region contains a set of voxels. We denote a voxel by $v_j$, where $j = 1, 2, \ldots, J$, and $J$ is the total number of voxels. We denote BOLD response obtained at each voxel $v_j$ by a function of time $t$ by $x_j(t)$, $\forall j$ and $\forall t$. Then, for each region $r_i$, we obtain a \textit{representative} time series $y_i(t)$, $\forall t$,  by spatially averaging the BOLD responses obtained from voxels residing in that region as follows:
\begin{equation}
\label{eq:average_sig}
y_i(t) = \dfrac{1}{J_i} \sum_{\forall v_j \in r_i} x_j(t) ,   \forall t,
\end{equation}
where $J_i$ is the number of voxels residing in the $i^{th}$ region $r_i$, $\forall i$.

\section{Encoding of Mesh Arc Descriptors}
In this section, we introduce our framework in order to encode MADs. First, we estimate MADs to represent the local connectivity patterns of cognitive tasks. Next, we reduce the dimension of MAD space using Principal Component Analysis (PCA). Then, we fit GMMs to the set of MADs. Finally, we encode MADs using three methods by employing their distance to the centers of GMMs, namely, Fisher Vectors (FV), Vector of Locally Aggregated Descriptors (VLAD) and Bag of Words (BoW). The encoded MADs are used to train an SVM classifier for cognitive task classification. Fig. ~\ref{fig:flowchart} represents the steps of our encoding framework.

\subsection{Estimation of Mesh Arc Descriptors (MADs)}

In this study, we form a star mesh around each region $r_i, \forall i$, by connecting the region to its functionally nearest neighboring regions. Functionally $p$-nearest neighbors of a region are defined by the regions having the $p$-largest Pearson correlation coefficient associated to that region during a cognitive task. 

We estimate the arc weights of each mesh formed around the region  $r_i$, in the $p$-nearest neighborhood  $\eta_p(r_i)$ of $r_i$ by assuming the following linear model;
\begin{equation}
	\label{eq:mesh}
	\mathbf{y}_i = \sum_{\forall r_j \in \eta_p(r_i)}{a_{i,j} \mathbf{y}_j + \lambda {a_{i,j}^2} + \mathbf{\epsilon}_i}     ,       \forall i,
\end{equation}
where $a_{i,j}, \forall i,j$ are estimated by minimizing the variance of error $\mathbf{\epsilon}_i$ using a ridge regression algorithm. In \eqref{eq:mesh}, $\lambda$ is the regularization parameter, and $\mathbf{y}_i$ is a vector of average BOLD response for region $r_i$, obtained during a cognitive task, such that $\mathbf{y}_i = [y_i(1), y_i(2), \ldots , y_i(T_c)]$, where $T_c$ denotes the number of time instances in fMRI measurements during task ${c \in [1,7]}$. The estimated arc weights $a_{i,j}$ represent the relationship between a region $r_i$ and its $p$-nearest neighbors $\eta_p(r_i)$ while a subject is exposed to a cognitive stimulus of length $T_c$. 
For each task, we solve \eqref{eq:mesh} for all regions $r_i$. Consequently, we obtain a MAD vector  $\mathbf{a}_i = [a_{i,1}, a_{i,2}, \ldots, a_{i,R}]$ of size $1 \times R$, for each cognitive task, such that
$a_{i,j} = 0 $ if $r_j \notin \eta_p(r_i)$.  Then, all the MADs are  concatenated  under a vector of size $1 \times R^2$, such that $\mathbf{a} = [\mathbf{a}_1, \mathbf{a}_2, \ldots, \mathbf{a}_R]$. 

\subsection{Encoding Methods}

In this subsection, we explain how MADs are encoded for fMRI data analysis.

\subsubsection{Fisher Vector (FV) Encoding}

Given a set of vectors, FV method encodes deviation of distribution of the vectors from a dictionary, which is typically described by a diagonal Gaussian mixture model (GMM) \citep{Carvajal2016}. In the proposed framework, our vectors are MADs obtained from fMRI data. We obtain a \textit{brain connectivity dictionary} of MADs by modeling the features using a GMM. In order to satisfy the assumption of diagonal covariance matrix of GMM and obtain linearly uncorrelated features, we apply PCA to MADs. For  simplicity, we use the same notation for MADs at the output of PCA, such that $\mathbf{a}_i$ represents the projection of a MADs in a $D$-dimensional space.

Let $A = \{\mathbf{a}_i \in \mathbb{R}^D \} _{i = 1} ^R$ denote a set of $D$-dimensional MADs which are obtained from a single task of a single subject, and are sampled from the set of all MADs, $\mathbb{A}$. Also, let $u_k$ be the $k^{th}$ component of GMM, which models the generative process for elements belonging to $\mathbb{A}$. We denote the parameter set of all $u_k$'s by $\lambda = \{ w_k, \bm{\mu}_k, \bm{\Sigma}_k \}_{k = 1} ^K$, where $w_k$, $\bm{\mu}_k$ and $\bm{\Sigma}_k$ are  the mixture weight, mean vector and covariance matrix of the $k^{th}$ Gaussian, respectively. We compute the mixture $u_\lambda$ by
$u_\lambda(\mathbf{a}_i) = \sum \limits_{k=1}^{K}w_k u_k(\mathbf{a}_i)$,
where $u_k(\mathbf{a}_i)$ denotes the $k^{th}$ Gaussian in the mixture, and is computed by
\begin{equation}
u_k(\mathbf{a}_i) = \dfrac{1}{(2\pi)^{D/2}|\bm{\Sigma}_k|^{1/2}}\exp \Big ( - \dfrac{1}{2} (\mathbf{a}_i - \bm{\mu}_k)^T \bm{\Sigma}_k^{-1} (\mathbf{a}_i - \bm{\mu}_k) \Big ).
\end{equation}

\begin{table*}[t!]
  \caption{Classification performance (\%) computed for different encoding methods.}
  \label{tab:classification}
  \centering
  \begin{tabular}{cccccccc}            
    \toprule
    $p$     & MAD     & FV-MAD & FV-MAD  & VLAD-MAD & VLAD-MAD & BoW-MAD & BoW-MAD\\
    &		&		  with PCA & no PCA & with PCA & no PCA &  with PCA & no PCA \\ 
    \midrule
    10 	   & 87.92 & \textbf{88.66} & 82.03 & 85.86 & 85.41 & 54.94 & 50.66 \\
    20     & 91.31 & \textbf{91.90} & 89.84 & 88.81 & 88.21 & 61.56 & 50.07 \\
    30     & 92.34 & \textbf{92.78} & 91.61 & 90.72 & 90.28 & 64.06 & 55.08 \\
    40     & 93.23 & \textbf{94.40} & 92.49 & 91.75 & 91.02 & 66.42 & 56.99 \\
    \bottomrule
  \end{tabular}
\end{table*}

We estimate the parameters of GMM on a training set of MADs using Expectation Maximization (EM) algorithm. We assume that diagonal covariance matrices are identified by $\bm{\sigma}^2_k$ which denotes a vector of the diagonal entries of $\bm{\Sigma}^2_k$.  Then, we compute the derivatives of log-likelihood of GMM with respect to the parameters using
$
\bm{G}_\lambda^A = R^{-1} \sum \limits _{i=1}^R \nabla_\lambda \log u_\lambda(\mathbf{a}_i).
$
The derivatives are only taken with respect to Gaussian means and variances, since the gradients computed with respect to the weight parameters, $w_i$, provide \textit{less information} \citep{Perronnin2010}. Therefore, we obtain a representation that captures the average first and second order differences between MADs and each of the GMM models.  Then, we compute $D$-dimensional gradients using
\begin{equation}
\bm{ \mathcal{G}}^A_{ \bm{\mu}_k}  = \dfrac{1}{R \sqrt{w_k}} \sum_{i=1}^R \gamma_i(k) \left( \dfrac{\mathbf{a}_i - \bm{\mu}_k }{ \bm{\sigma}_k } \right) \; {\rm }
\end{equation}
and
\begin{equation}
\bm{ \mathcal{G}}^A_{\bm{\sigma}_k}  = \dfrac{1}{R \sqrt{2 w_k}} \sum_{i=1}^R \gamma_i(k) \left[ \dfrac{(\mathbf{a}_i - \bm{\mu}_k )^2}{ \bm{ \sigma}_k ^2} -1 \right],
\end{equation}
where $\gamma_i(k)$ represents the soft assignment of the MAD vector $\mathbf{a}_i$ to the $k^{th}$ Gaussian, and is defined by $
{\gamma_i(k) = w_k u_k(\mathbf{a}_i) \Big (\sum \limits_{l=1}^K w_l u_l(\mathbf{a}_i) \Big )^{-1}}$. 

A Fisher vector $\bm{G}_\lambda^A$ is obtained by concatenating the gradients under a vector   ${\bm{G}_\lambda^A = \left[ \bm{ \mathcal{G}}^A_{\bm{\mu}_1} , \bm{ \mathcal{G}}^A_{\sigma_1} , \bm{ \mathcal{G}}^A_{\bm{\mu}_2} , \bm{ \mathcal{G}}^A_{\bm{\sigma}_2}, \ldots, \bm{ \mathcal{G}}^A_{\bm{\mu}_K} , \bm{ \mathcal{G}}^A_{\bm{\sigma}_K}   \right]}$ of size $2 K D$. We obtain Fisher Vectors for both training and test data. In order to obtain better accuracy, we further apply $l_2$ normalization and square-root transformation on FVs as suggested in \citep{Sanchez2013}.

\subsubsection{Vector of Locally Aggregated Descriptors (VLAD)}

VLAD encodes a set of descriptors into a dictionary, which is computed by k-means clustering method. In the proposed framework, we first perform k-means clustering of MADs on the training data. Hence, the cluster centroids $\{\bm{\mu} \}_{k=1}^K$ correspond to our codewords of \textit{brain connectivity dictionary}. Then, we associate each MAD $\mathbf{a}_i$ to its nearest codeword $NN(\mathbf{a}_i)$ in the dictionary, such that $NN(\mathbf{a}_i) = \underset{\bm{\mu}_k}{\mathrm{argmin}} \| \mathbf{a}_i - \bm{\mu}_k  \|_2 $, where $\| \cdot \|_2$ is the Euclidean norm. 

Recall from the previous subsection that, $A = \{\mathbf{a}_i\}_{i = 1}^R$ represents the set of $D$-dimensional MADs belonging to a single task of a single subject, each of which is sampled from $\mathbb{A}$ that contains all MADs belonging to all tasks of all subjects. For each codeword $\bm{\mu}_k$, we compute the sum of the differences, $(\mathbf{a}_i - \bm{\mu}_k)$, of the descriptors $\mathbf{a}_i$, which are assigned to the $k^{th}$ cluster by
\begin{equation}
\bm{ \mathfrak{v}}_k^A = \sum_{\mathbf{a}_i : NN(\mathbf{a}_i) = \bm{\mu}_k} (\mathbf{a}_i - \bm{ \mu}_k) .
\end{equation}
We concatenate the $D$-dimensional vectors $\bm{\mathfrak{v}}_k^A$ for all clusters, and obtain a $KD$ dimensional VLAD encoding ${\bm{ \mathcal{V}}^A  = [ \bm{ \mathfrak{v}}_1^A , \bm{ \mathfrak{v}}_2^A, \ldots, \bm{ \mathfrak{v}}_K^A]}$. Note that, we obtain VLAD encoding for each $A$, computed using training and test data.

\subsubsection{Bag-of-Words (BoW)}
 We also encode MADs using the BoW approach, where the words correspond to a set of selected MADs. We first cluster MADs obtained from training data using a k-means clustering method. Then, for each $A$, we compute the number of MADs, $n^A_k$, belonging to the $k^{th}$ cluster. Finally, we represent each sample by a $K$-dimensional vector such that $\bm{ N}^A  = [n^A_1, n^A_2, \ldots, n^A_k]$.

\begin{figure*}[t!]
	\centering
    \begin{subfigure}[t]{0.35\textwidth}
		\includegraphics[width=\textwidth]{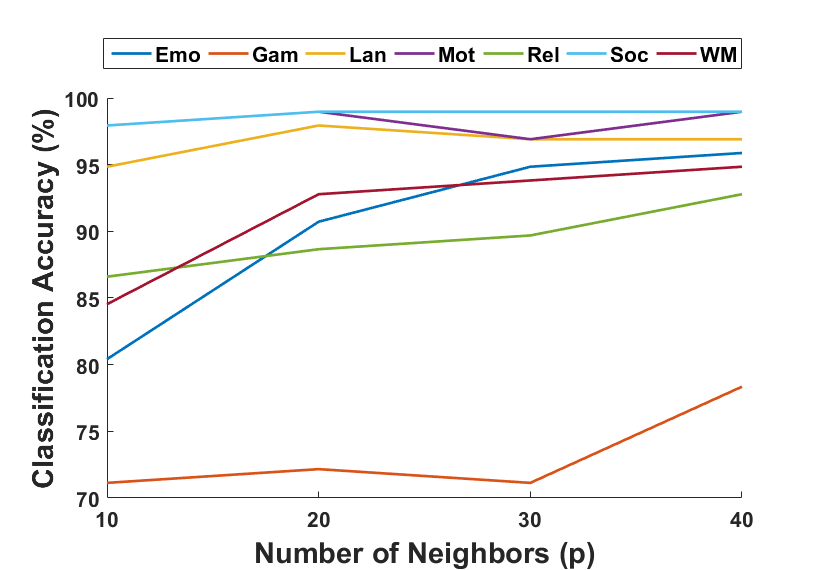}
        \caption{MAD}
		\label{fig:confmat_mad}
	\end{subfigure}	\hspace{3em}
	\begin{subfigure}[t]{0.35\textwidth}
		\includegraphics[width=\textwidth]{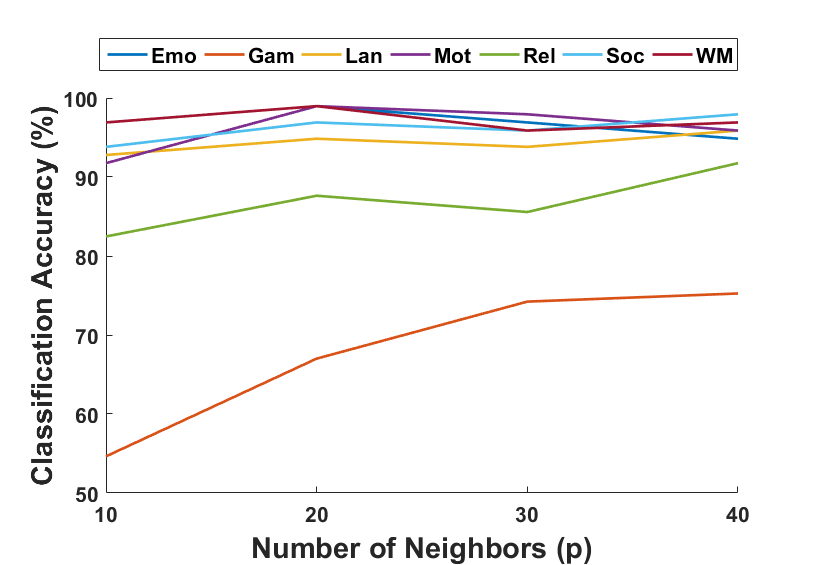}
        \caption{FV-MAD}
		\label{fig:confmat_fv}
	\end{subfigure}	
    
    \begin{subfigure}[t]{0.35\textwidth}
		\includegraphics[width=\textwidth]{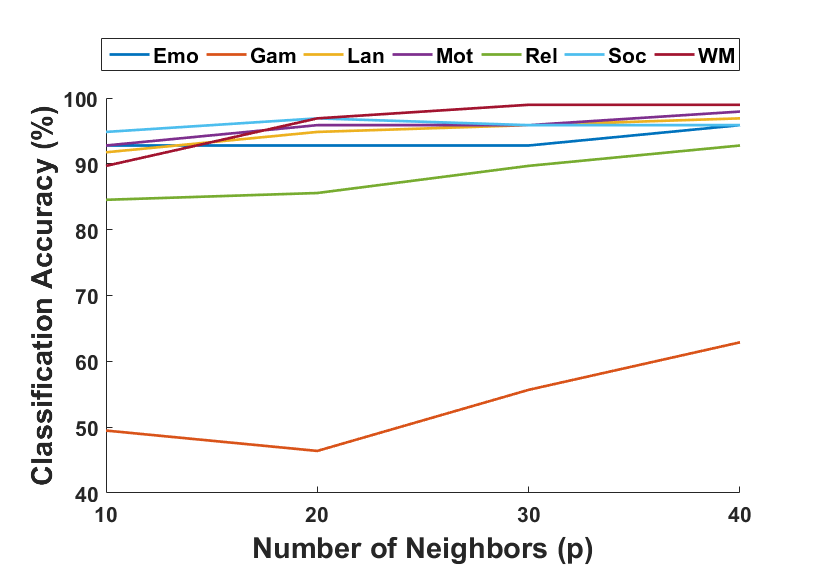}
        \caption{VLAD-MAD}
        \label{fig:confmat_vlad}
	\end{subfigure}	\hspace{3em}
    \begin{subfigure}[t]{0.35\textwidth}
		\includegraphics[width=\textwidth]{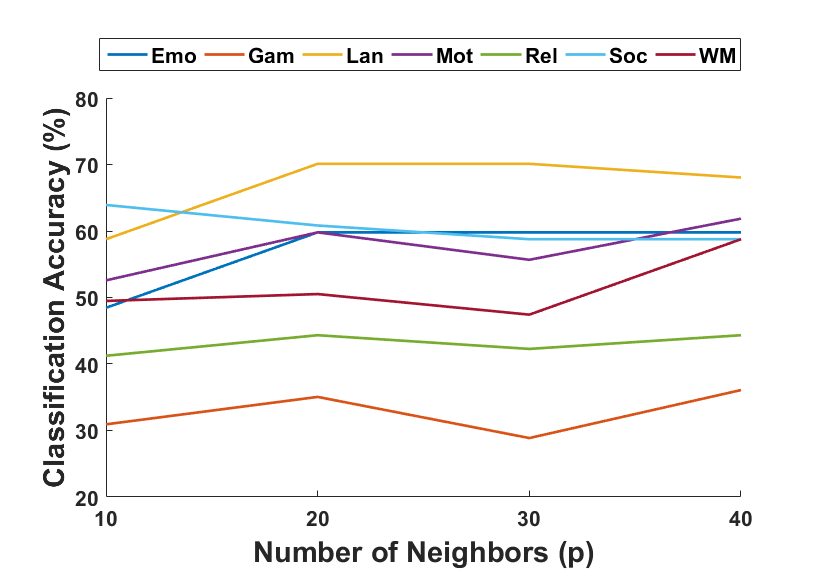}
        \caption{BoW-MAD}
        \label{fig:confmat_bow}
	\end{subfigure}	
    \caption{Classification accuracy (\%) measured for each task.}
    \label{fig:confmats}
\end{figure*}

\section{Experimental Results}
We perform two sets of experiments. First, we classify the cognitive tasks of HCP datasets using the raw and encoded MADs to see the effect of encoding on MADs and compare the performance of various encoding methods. In order to examine the power of MADs compared to popular fMRI representation methods, we also classify the cognitive tasks by using average BOLD responses and their Pearson correlations. Finally, we visualize and analyze the codewords of the proposed \textit{brain connectivity dictionary}.

\subsection{Classification Results}

In order to perform classification, we first compute FV ($\bm{ G}_\lambda^A $), VLAD ($\bm{ \mathcal{V}}^A $) and BoW ($\bm{ N}^A $) encoding of local MADs for all tasks and all participants. Then, we use the encoded MAD vectors to train and test linear SVM classifiers. We measure the classification performance using a 10-fold cross validation (CV) scheme by randomly splitting the data into 10 subsets  according to a uniform distribution, training each model on 9 splits, and testing the learned model on the remaining split. Note that, training and test splits contain data associated to different participants for each fold. In the proposed framework, we use only training data for implementation  of all methods that are employed in learning phase, such as computation of PCA using MADs, estimation of Gaussian mixture models, k-means clustering and training SVM classifiers. We examine the proposed methods for various number of neighbors $p \in P =\{10,20,30,40\}$, in each mesh to observe the effect of degree of locality of the meshes. Considering the fact that MADs are $R$-dimensional vectors (before application of PCA), employment of smaller $p$ values provides increased locality and sparser MAD vectors. Before employment of FV encoding, we project MADs onto $d \in \{50, 60, 70, 80, 90\}$ dimensional spaces. Moreover, we select the number of Gaussians in GMM and the number of clusters in k-means using $k \in \{20, 40, 60, 80, 100, 120\}$. We compute average performance of 10-fold CV for all $k$ and $d$. Accuracy values denoted in Table \ref{tab:classification} show maximum of these performances with respect to the values of $k$ and $d$.

Results given in Table \ref{tab:classification} show that, FV encoding of MADs computed by employment of PCA (FV-MAD with PCA) gives the best performance to classify cognitive tasks compared to VLAD (VLAD-MAD) and  BoW encoding (BoW-MAD). 
We observe that VLAD encoding of MADs does not increase the discrimination power of MADs. Although accuracy values of VLAD-MAD are comparable with that of MAD, BoW representations of MADs yield substantial decrease in accuracy. In order to observe the representation power of MAD compared to popular fMRI representation methods, we use average voxel BOLD responses and Pearson correlation pairs between them as the input feature vectors of SVM classifiers. When the average region BOLD responses are employed for classification, we obtain  almost chance accuracy $(14.29\%)$. On the other hand, when pairwise correlations between all pairs of region BOLD responses are used as features, we obtain a classification accuracy of $77.49\%$.  In Table \ref{tab:comparison_accuracy}, we observe a substantial increase $(10-15\%)$ obtained by representing fMRI data by MADs. These observations show the power of MADs for representation of the connectivity of anatomic regions.

\begin{table}[]
\centering
\caption{Classification accuracy (\%) measured with different descriptors.}
\label{tab:comparison_accuracy}
\begin{tabular}{ccc}
\textbf{BOLD Response}  & \textbf{Pearson Correlation }& \textbf{MAD}           \\ \hline
14.29 & 77.49         & 87.92 - 93.23 \\ \hline
\end{tabular}
\end{table}

\begin{figure}[ht!]
	\centering
	\includegraphics[scale=0.5]{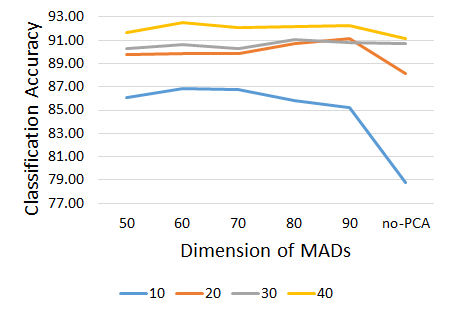}
	\caption{Analysis of effect of dimension reduction using PCA on FV to classification accuracy for different number of neighbors $p \in P$.  }
	\label{fig:PCA_figure}
\end{figure}

\begin{figure*}[ht!]
	\captionsetup[subfigure]{labelformat=empty}	
	\centering
    \begin{subfigure}[t]{0.12\textwidth}
		\includegraphics[width=\textwidth]{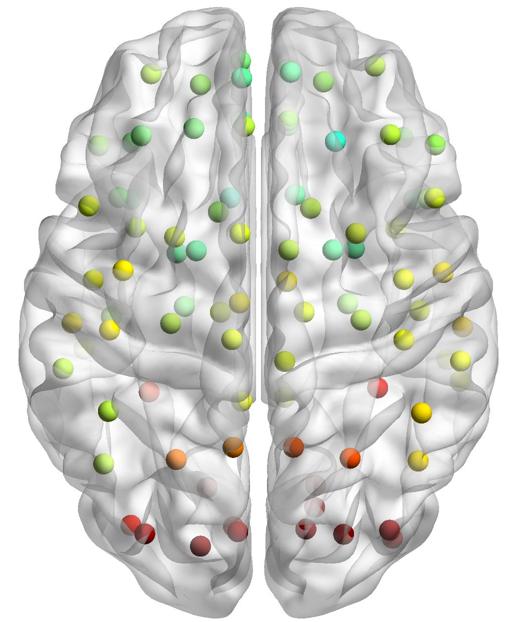}
        \caption{Codeword1}
	\end{subfigure}	\hspace{.7em}
	\begin{subfigure}[t]{0.12\textwidth}
		\includegraphics[width=\textwidth]{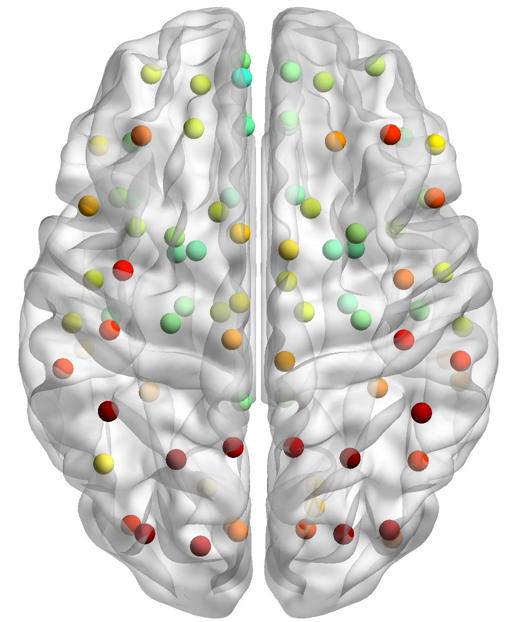}
        \caption{Codeword2}
	\end{subfigure}	\hspace{.7em}
    \begin{subfigure}[t]{0.12\textwidth}
		\includegraphics[width=\textwidth]{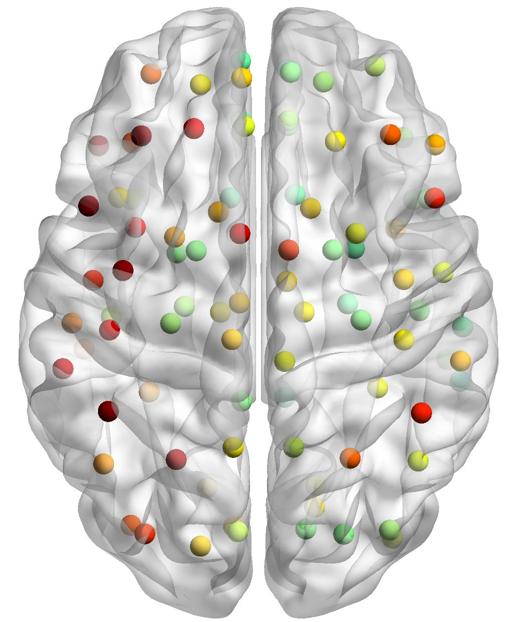}
        \caption{Codeword3}
	\end{subfigure} \hspace{.7em}
    \begin{subfigure}[t]{0.12\textwidth}
		\includegraphics[width=\textwidth]{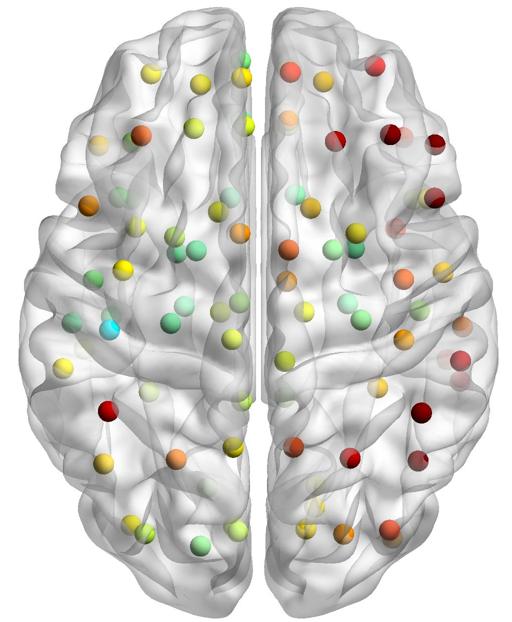}
        \caption{Codeword4}
	\end{subfigure}	\hspace{.7em}
    \begin{subfigure}[t]{0.12\textwidth}
		\includegraphics[width=\textwidth]{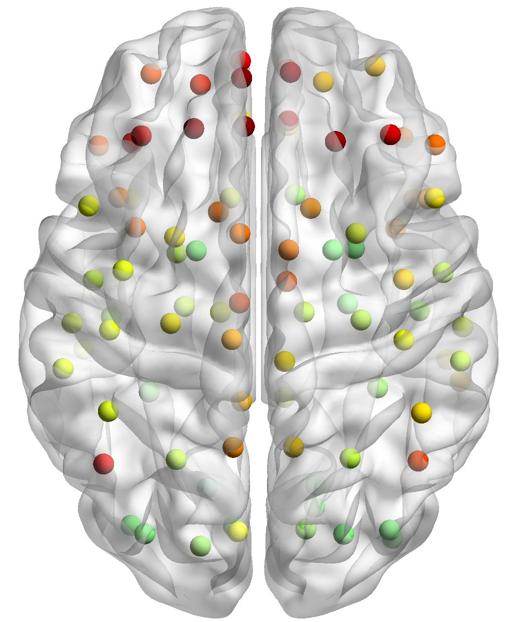}
        \caption{Codeword5}
	\end{subfigure}	  \hspace{.7em}
    \begin{subfigure}[t]{0.12\textwidth}
		\includegraphics[width=\textwidth]{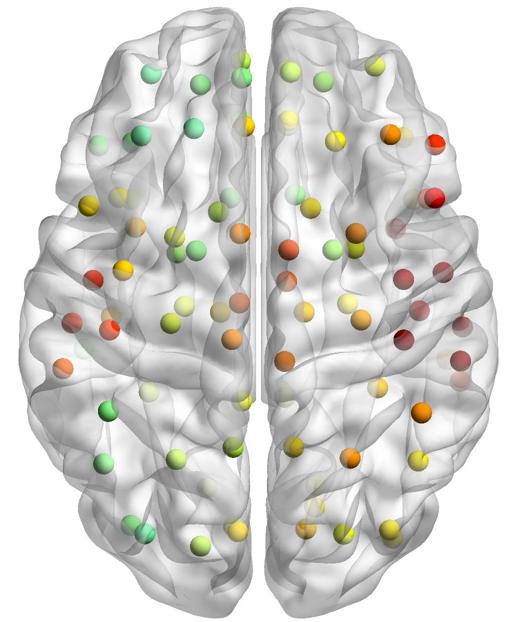}
        \caption{Codeword6}
	\end{subfigure}	\hspace{.7em}
    \begin{subfigure}[t]{0.12\textwidth}
		\includegraphics[width=\textwidth]{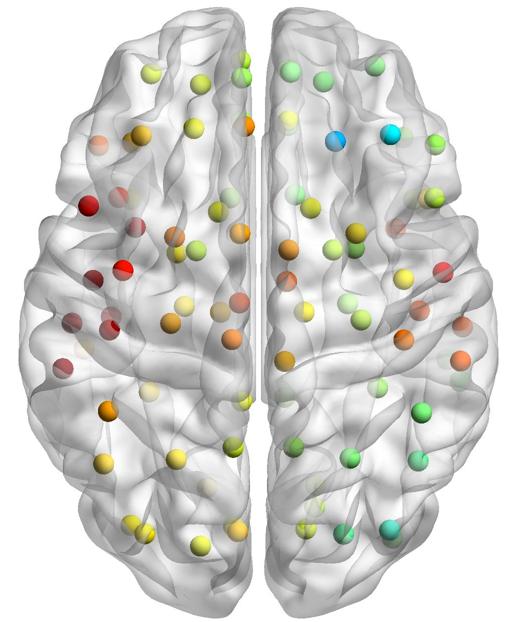}
        \caption{Codeword7}
	\end{subfigure}	 
    
    \begin{subfigure}[t]{0.12\textwidth}
		\includegraphics[width=\textwidth]{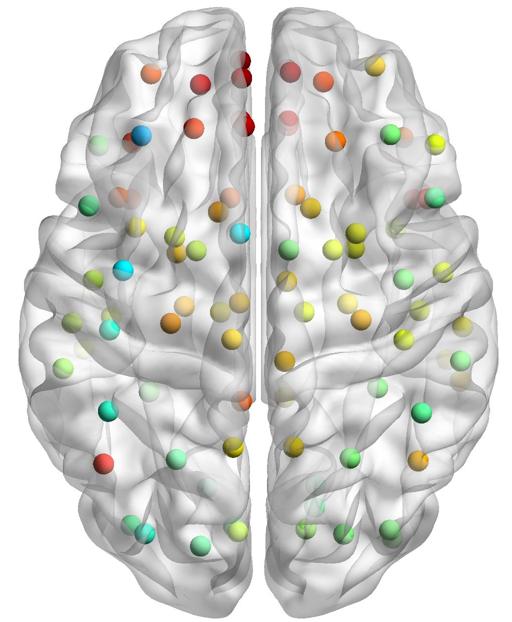}
        \caption{Codeword8}
	\end{subfigure}	\hspace{.7em}
    \begin{subfigure}[t]{0.12\textwidth}
		\includegraphics[width=\textwidth]{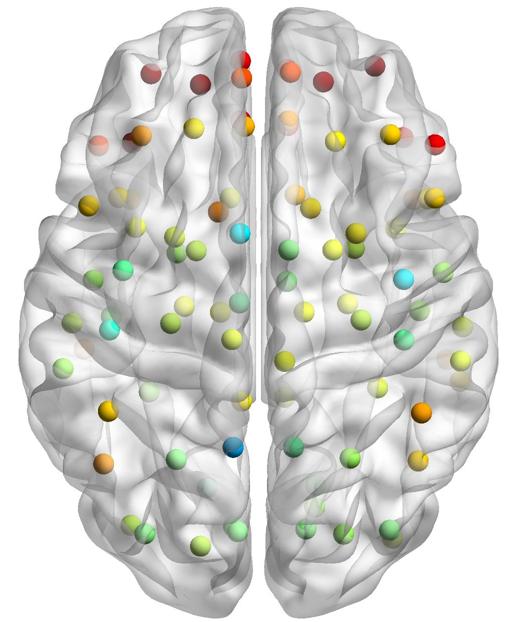}
        \caption{Codeword9}
	\end{subfigure}	\hspace{.7em}
    \begin{subfigure}[t]{0.12\textwidth}
		\includegraphics[width=\textwidth]{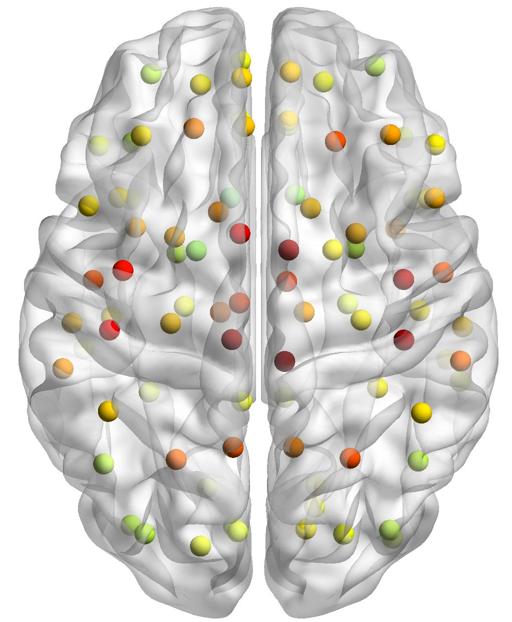}
        \caption{Codeword10}
	\end{subfigure}\hspace{.7em}    
	\begin{subfigure}[t]{0.12\textwidth}
		\includegraphics[width=\textwidth]{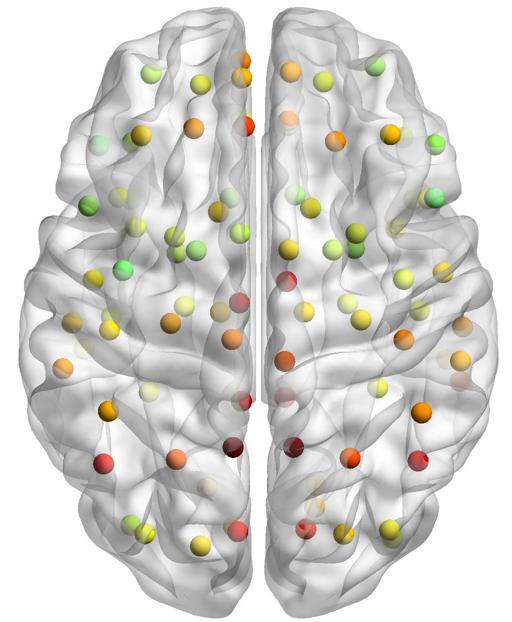}
		\caption{Codeword11}
	\end{subfigure}\hspace{.7em}
    \begin{subfigure}[t]{0.12\textwidth}
		\includegraphics[width=\textwidth]{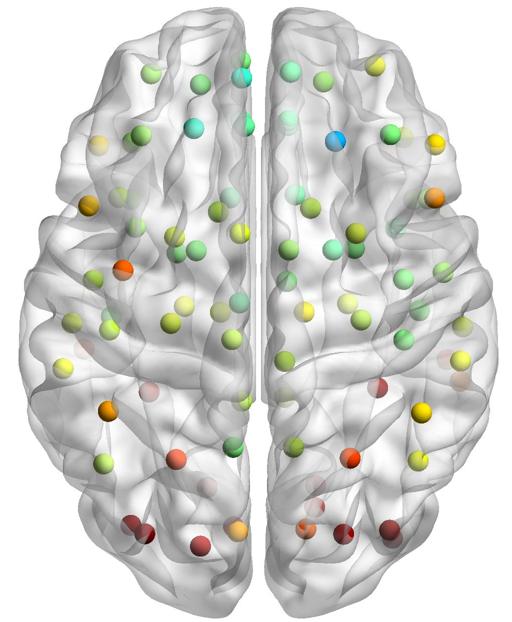}
        \caption{Codeword12}
	\end{subfigure}	\hspace{.7em}
    \begin{subfigure}[t]{0.12\textwidth}
		\includegraphics[width=\textwidth]{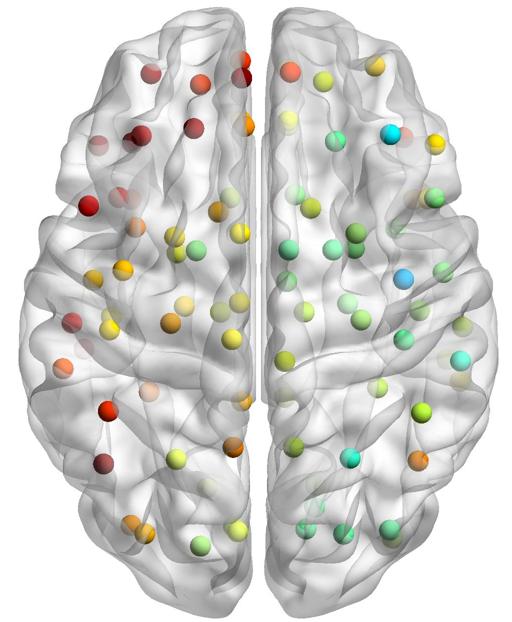}
        \caption{Codeword13}
	\end{subfigure}\hspace{.7em}
    \begin{subfigure}[t]{0.12\textwidth}
		\includegraphics[width=\textwidth]{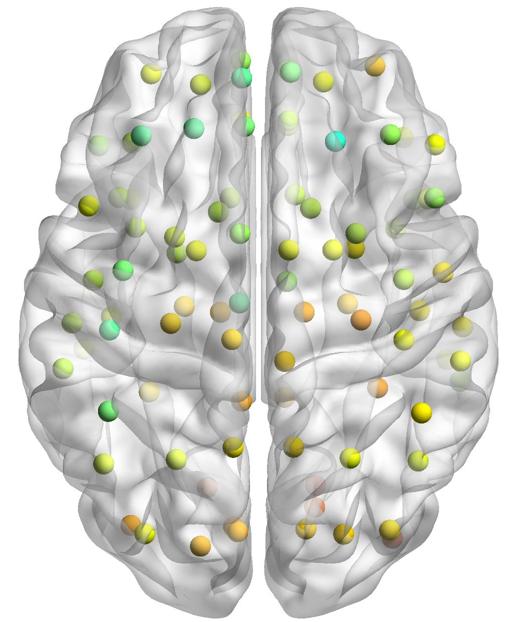}
        \caption{Codeword14}
	\end{subfigure}	
    
    \begin{subfigure}[t]{0.12\textwidth}
		\includegraphics[width=\textwidth]{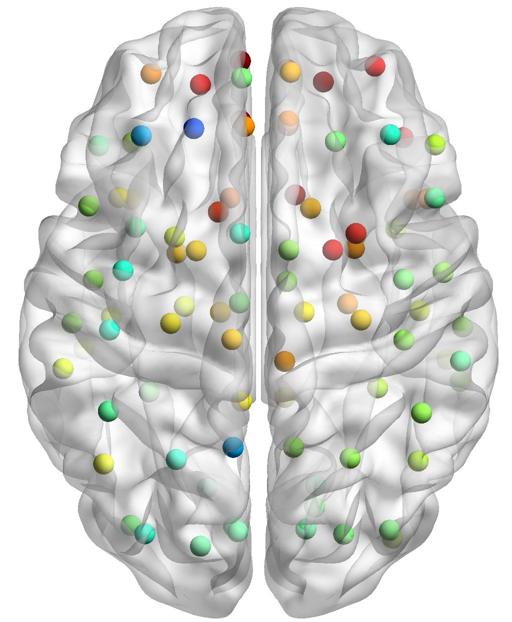}
        \caption{Codeword15}
	\end{subfigure}	\hspace{.7em}  
    \begin{subfigure}[t]{0.12\textwidth}
		\includegraphics[width=\textwidth]{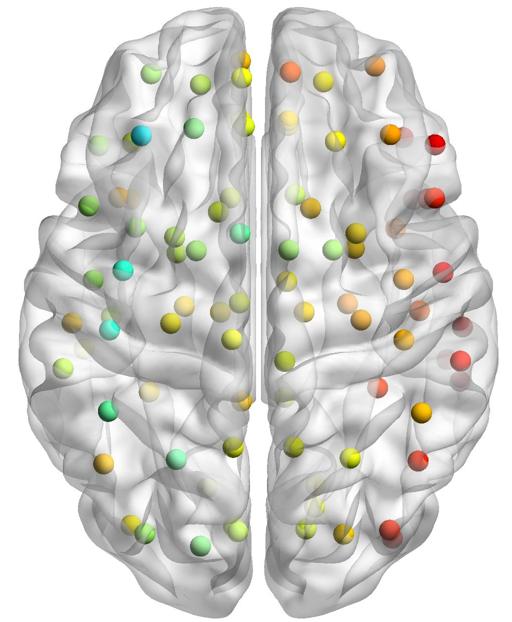}
		\caption{Codeword16}
	\end{subfigure}	\hspace{.7em}
    \begin{subfigure}[t]{0.12\textwidth}
		\includegraphics[width=\textwidth]{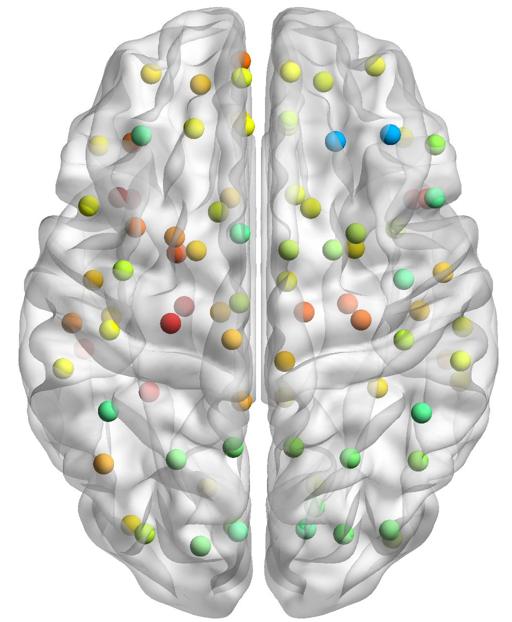}
        \caption{Codeword17}
	\end{subfigure}	\hspace{.7em}
    \begin{subfigure}[t]{0.12\textwidth}
		\includegraphics[width=\textwidth]{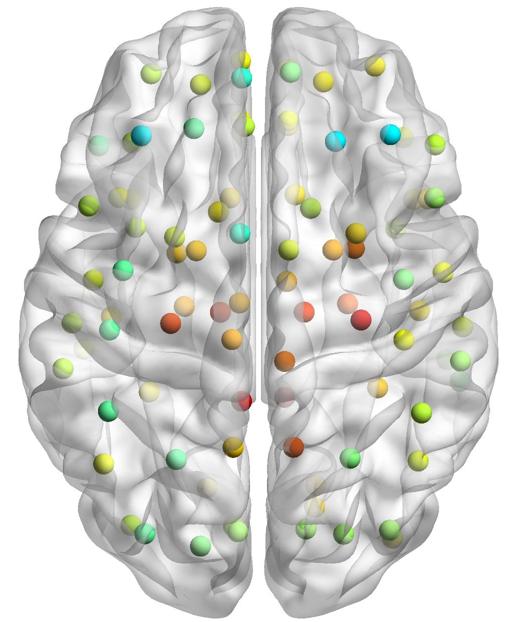}
        \caption{Codeword18}
	\end{subfigure}\hspace{.7em}  
    \begin{subfigure}[t]{0.12\textwidth}
		\includegraphics[width=\textwidth]{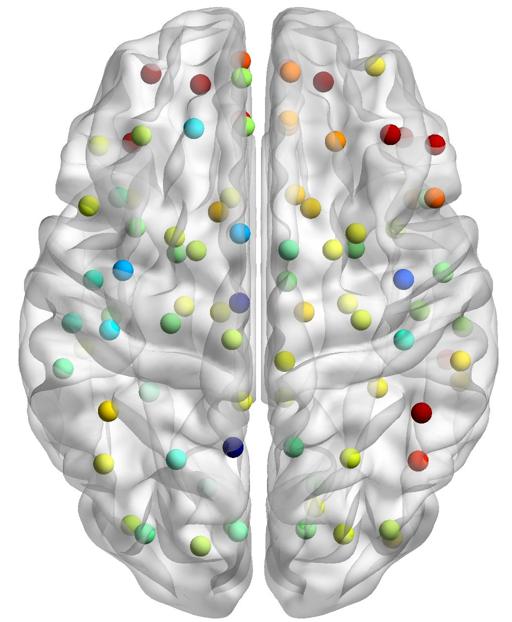}
        \caption{Codeword19}
	\end{subfigure}	\hspace{.7em}
    \begin{subfigure}[t]{0.12\textwidth}
		\includegraphics[width=\textwidth]{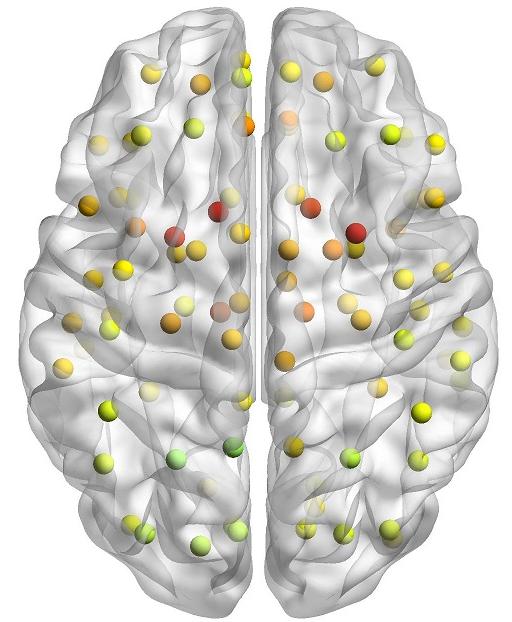}
        \caption{Codeword20}
	\end{subfigure}	\hspace{.7em}
    \begin{subfigure}[t]{0.12\textwidth}
		\includegraphics[width=\textwidth]{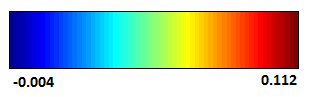}
	\end{subfigure}	
    
	\caption{Codewods of brain connectivity dictionary  obtained by FVs for encoding of MADs with $p= 40$. Codewords are visualized using BrainNet Viewer \citep{Xia2013}, and depicted in an order such that Codeword1 and Codeword20 represent the codeword of GHE, and GLE, respectively. }
    \label{fig:codewords}
    
\end{figure*}

Next, we examine the success of encoding for each cognitive task. We observe in Fig.~\ref{fig:confmats} that the accuracy values obtained for Gambling and Relational Processing tasks are  lower compared to the other tasks performed using FV-MAD and VLAD-MAD. However, accuracy values obtained from BoW-MAD for all classes are low, meaning that BoW is not successful to encode MADs. When we compare the results shown in Fig.~\ref{fig:confmat_mad},  Fig.~\ref{fig:confmat_fv} and Fig.~\ref{fig:confmat_vlad}, we observe that we can boost the accuracy for Emotion and WM tasks compared to concatenated MADs by encoding  MADs with FV and VLAD. Yet, the accuracy is decreased for Gambling task, especially for the lower values of $p$.

We also analyze the effect of employment of PCA to local MADs before computation of FV encoding. Our results verify the observation that dimension reduction using PCA is a key method used to boost the performance (see Table \ref{tab:classification}). We also observe from Table \ref{tab:classification} that without PCA, FV encoding cannot improve the performance of concatenated MADs for any value of $p$. Similarly, we observe that performance decreases if we do not employ PCA for MADs using VLAD and BoW. We examine how accuracy changes as dimension of MAD is decreased using PCA in Fig.~\ref{fig:PCA_figure} by plotting the average accuracy obtained over all values of $k$ for different $d \in \{ 50,60,\ldots,90 \}$. We observe that, if the dimension $d$ of the low-dimensional space obtained using PCA increases, then average accuracy may increase. Moreover, without dimension reduction, the average accuracy decreases for most of the $p \in P$ values.

\subsection{Analysis of Codewords }

In this subsection, we explore the contribution of a particular codeword (component of GMM) to classify the cognitive tasks  by measuring the energy (Euclidean norm) of the corresponding columns of FVs. Recall that, the  columns of the $k^{th}$ Gaussian in a FV encoding is $ [ \bm{\mathcal{G}}^A_{\bm{\mu}_K}, \bm{ \mathcal{G}}^A_{ \bm{\sigma}_K} ]$  for a set $A$ of MADs. If we denote the corresponding columns of the $k^{th}$ Gaussian in the FV encoding using all training data by $ \bm{\mathcal{G}}_k =  [ \bm{\mathcal{G}}_{\bm{\mu}_K} , \bm{ \mathcal{G}}_{\bm{\sigma}_K} ]$, then the energy is measured by  the Euclidean norm $ \| \bm{\mathcal{G}}_k \|_2 $. We denote a Gaussian whose FV columns have the lowest energy by GLE, and the one whose FV columns have the highest energy by GHE.

Next, we visualize $R$-dimensional codewords obtained from MADs without using PCA. Each codeword corresponds to a mesh pattern formed around a seed voxel. Large values of codeword elements represent that, a MAD formed between the region corresponding to the codeword element and seed region has large value. Similarly, small values of codeword elements indicate small values of MADs between seed region and the corresponding region. As an example, we plot codewords for $p = 40$ and $k = 20$, in Fig.~\ref{fig:codewords}. Notice from Figure \ref{fig:codewords} that, codewords represent different mesh patterns. For example, Codeword1 represents a mesh pattern in which MADs formed between a seed region and regions residing in the occipital lobe have large values while MADs formed between a seed region and the other regions have smaller values. Recall that, each GMM component obtained without using PCA corresponds to $2R$ columns of FVs. In Fig.~\ref{fig:codewords}, we provide visualization of the Gaussians in an order from GHE to GLE by plotting the codewords of the corresponding Gaussians. The results show that codeword of GHE has large values mainly in the occipital lobe. On the other hand, codeword of GLE has large values in central structures including Caudate, Putamen and Thalamus.

\begin{figure}[h!]
	\centering
    \vspace{-0.3cm}
		\includegraphics[width=.42\textwidth]{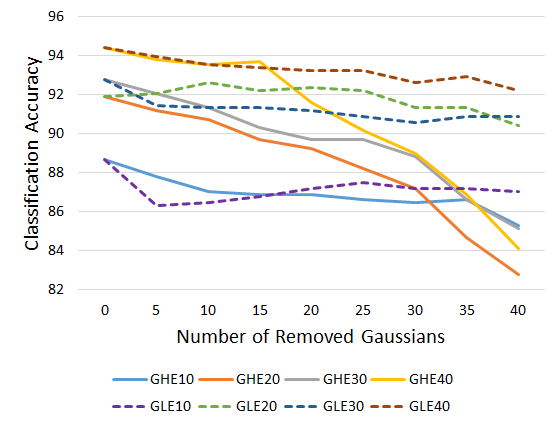}	
    \caption{Analysis of effect of selection and removal of FV encoding of Gaussians to classification accuracy.}
    \label{fig:removed}
    \vspace{-0.3cm}
\end{figure}

\cite{Simonyan2013} showed that GHEs represent the facial features whereas the GLEs cover the background areas. In other words, they stated that GHEs are significant for comparing human face images. Following our results, we conjecture that the GHEs consisting of MADs convey class (task) discriminative information. In order to validate this assumption, and investigate the relationship between the energy of FV columns related to Gaussians and the classification accuracy, we employ an energy based feature selection scheme for FV encoding. For this purpose, we select and remove the corresponding columns from FVs, and perform classification. When we remove FV columns with minimum and maximum energy, we observe that the accuracy decreases slightly and significantly, respectively (see Fig.~\ref{fig:removed}). We conclude that, GHEs provide more class (task) discrimination power to classifiers compared to GLEs.

\section{Discussion and Conclusion}

In this study, we propose a novel framework to encode the local connectivity patterns of fMRI data for classification of the cognitive states of the HCP task fMRI data set. fMRI connectivity patterns are modeled by an ensemble of local meshes formed around each anatomic region. MADs are estimated by using ridge regression, assuming a linear relationship among anatomic regions. Several state-of-the-art methods, such as FV, VLAD and BoW, are used to encode MADs. We, also, generate a \textit{brain connectivity dictionary} by fitting a GMM to the set of all MADs to analyze the connectivity patterns of fMRI data with respect to anatomical regions. 

We observe that FV encoding of MADs is more successful compared to VLAD and BoW methods. Moreover, FV encoding improves the performance of models that employ MADs. We also observe that statistical decorrelation of MADs obtained using PCA is a significant step for FV encoding. We further analyze class (task) discrimination  power of Gaussians obtained from FV encoding. We observe a relationship between the energy of FV columns corresponding to Gaussians and classification accuracy. Following that result, we visualize codewords of \textit{brain connectivity dictionary} on a human brain template according to their discriminative power, which is measured by an energy function. 

In this work, we constructed mesh structures of anatomical regions. As a future work, we plan to employ encoding of MADs by constructing mesh structures among the voxels in high dimensional spaces. In addition, we plan to employ the proposed framework to further analyze the relationship between different tasks and cognitive processes by visualizing feature maps of MADs and their encoding on the AAL brain atlas.

\section*{Acknowledgments}
This work was supported by CREST, JST, and TUBITAK Project No 114E045. Itir Onal Ertugrul was supported by TUBITAK.

\bibliographystyle{model2-names}
\bibliography{refs}

\begin{thebibliography}{27}
\expandafter\ifx\csname natexlab\endcsname\relax\def\natexlab#1{#1}\fi
\providecommand{\url}[1]{\texttt{#1}}
\providecommand{\href}[2]{#2}
\providecommand{\path}[1]{#1}
\providecommand{\DOIprefix}{doi:}
\providecommand{\ArXivprefix}{arXiv:}
\providecommand{\URLprefix}{URL: }
\providecommand{\Pubmedprefix}{pmid:}
\providecommand{\doi}[1]{\href{http://dx.doi.org/#1}{\path{#1}}}
\providecommand{\Pubmed}[1]{\href{pmid:#1}{\path{#1}}}
\providecommand{\bibinfo}[2]{#2}
\ifx\xfnm\relax \def\xfnm[#1]{\unskip,\space#1}\fi
\bibitem[{Abbas et~al.(2015)Abbas, Deligiannis and Andreopoulos}]{Abbas2015}
\bibinfo{author}{Abbas, A.}, \bibinfo{author}{Deligiannis, N.},
  \bibinfo{author}{Andreopoulos, Y.}, \bibinfo{year}{2015}.
\newblock \bibinfo{title}{Vectors of locally aggregated centers for compact
  video representation}, in: \bibinfo{booktitle}{IEEE ICME}, pp.
  \bibinfo{pages}{1--6}.
\bibitem[{Barch et~al.(2013)Barch, Burgess, Harms, Petersen, Schlaggar,
  Corbetta, Glasser, Curtiss, Dixit, Feldt et~al.}]{Barch2013}
\bibinfo{author}{Barch, D.M.}, \bibinfo{author}{Burgess, G.C.},
  \bibinfo{author}{Harms, M.P.}, \bibinfo{author}{Petersen, S.E.},
  \bibinfo{author}{Schlaggar, B.L.}, \bibinfo{author}{Corbetta, M.},
  \bibinfo{author}{Glasser, M.F.}, \bibinfo{author}{Curtiss, S.},
  \bibinfo{author}{Dixit, S.}, \bibinfo{author}{Feldt, C.}, et~al.,
  \bibinfo{year}{2013}.
\newblock \bibinfo{title}{Function in the human connectome: task-fmri and
  individual differences in behavior}.
\newblock \bibinfo{journal}{Neuroimage} \bibinfo{volume}{80},
  \bibinfo{pages}{169--189}.
\bibitem[{Carvajal et~al.(2016)Carvajal, McCool, Lovell and
  Sanderson}]{Carvajal2016}
\bibinfo{author}{Carvajal, J.}, \bibinfo{author}{McCool, C.},
  \bibinfo{author}{Lovell, B.}, \bibinfo{author}{Sanderson, C.},
  \bibinfo{year}{2016}.
\newblock \bibinfo{title}{Joint recognition and segmentation of actions via
  probabilistic integration of spatio-temporal fisher vectors}.
\newblock \bibinfo{journal}{arXiv preprint arXiv:1602.01601} .
\bibitem[{Delhumeau et~al.(2013)Delhumeau, Gosselin, J{\'e}gou and
  P{\'e}rez}]{Delhumeau2013}
\bibinfo{author}{Delhumeau, J.}, \bibinfo{author}{Gosselin, P.H.},
  \bibinfo{author}{J{\'e}gou, H.}, \bibinfo{author}{P{\'e}rez, P.},
  \bibinfo{year}{2013}.
\newblock \bibinfo{title}{Revisiting the vlad image representation}, in:
  \bibinfo{booktitle}{ACMMM}, pp. \bibinfo{pages}{653--656}.
\bibitem[{Haxby et~al.(2001)Haxby, Gobbini, Furey, Ishai, Schouten and
  Pietrini}]{Haxby2001}
\bibinfo{author}{Haxby, J.V.}, \bibinfo{author}{Gobbini, M.I.},
  \bibinfo{author}{Furey, M.L.}, \bibinfo{author}{Ishai, A.},
  \bibinfo{author}{Schouten, J.L.}, \bibinfo{author}{Pietrini, P.},
  \bibinfo{year}{2001}.
\newblock \bibinfo{title}{Distributed and overlapping representations of faces
  and objects in ventral temporal cortex}.
\newblock \bibinfo{journal}{Science} \bibinfo{volume}{293},
  \bibinfo{pages}{2425 -- 2429}.
\bibitem[{Jaakkola et~al.(1999)Jaakkola, Haussler et~al.}]{Jaakkola1999}
\bibinfo{author}{Jaakkola, T.S.}, \bibinfo{author}{Haussler, D.}, et~al.,
  \bibinfo{year}{1999}.
\newblock \bibinfo{title}{Exploiting generative models in discriminative
  classifiers}.
\newblock \bibinfo{journal}{NIPS} , \bibinfo{pages}{487--493}.
\bibitem[{J{\'e}gou et~al.(2010)J{\'e}gou, Douze, Schmid and
  P{\'e}rez}]{Jegou2010}
\bibinfo{author}{J{\'e}gou, H.}, \bibinfo{author}{Douze, M.},
  \bibinfo{author}{Schmid, C.}, \bibinfo{author}{P{\'e}rez, P.},
  \bibinfo{year}{2010}.
\newblock \bibinfo{title}{Aggregating local descriptors into a compact image
  representation}, in: \bibinfo{booktitle}{CVPR}, pp.
  \bibinfo{pages}{3304--3311}.
\bibitem[{Liu et~al.(2014)Liu, Shen, Wang, van~den Hengel and Wang}]{Liu2014}
\bibinfo{author}{Liu, L.}, \bibinfo{author}{Shen, C.}, \bibinfo{author}{Wang,
  L.}, \bibinfo{author}{van~den Hengel, A.}, \bibinfo{author}{Wang, C.},
  \bibinfo{year}{2014}.
\newblock \bibinfo{title}{Encoding high dimensional local features by sparse
  coding based fisher vectors}, in: \bibinfo{booktitle}{NIPS}, pp.
  \bibinfo{pages}{1143--1151}.
\bibitem[{Merino et~al.(2013)Merino, Meng, Gordon, Lance, Johnson, Paul,
  Robbins, Vettel and Huang}]{Merino2013}
\bibinfo{author}{Merino, L.M.}, \bibinfo{author}{Meng, J.},
  \bibinfo{author}{Gordon, S.}, \bibinfo{author}{Lance, B.J.},
  \bibinfo{author}{Johnson, T.}, \bibinfo{author}{Paul, V.},
  \bibinfo{author}{Robbins, K.}, \bibinfo{author}{Vettel, J.M.},
  \bibinfo{author}{Huang, Y.}, \bibinfo{year}{2013}.
\newblock \bibinfo{title}{A bag-of-words model for task-load prediction from
  eeg in complex environments}, in: \bibinfo{booktitle}{ICASSP}, pp.
  \bibinfo{pages}{1227--1231}.
\bibitem[{Mitchell et~al.(2004)Mitchell, Hutchinson, Niculescu, Pereira, Wang,
  Just and Newman}]{mitchell2004}
\bibinfo{author}{Mitchell, T.M.}, \bibinfo{author}{Hutchinson, R.},
  \bibinfo{author}{Niculescu, R.S.}, \bibinfo{author}{Pereira, F.},
  \bibinfo{author}{Wang, X.}, \bibinfo{author}{Just, M.},
  \bibinfo{author}{Newman, S.}, \bibinfo{year}{2004}.
\newblock \bibinfo{title}{Learning to decode cognitive states from brain
  images}.
\newblock \bibinfo{journal}{Machine Learning} \bibinfo{volume}{57},
  \bibinfo{pages}{145--175}.
\bibitem[{Onal et~al.(2015a)Onal, Ozay and Yarman~Vural}]{Onal2015_1}
\bibinfo{author}{Onal, I.}, \bibinfo{author}{Ozay, M.},
  \bibinfo{author}{Yarman~Vural, F.}, \bibinfo{year}{2015}a.
\newblock \bibinfo{title}{Modeling voxel connectivity for brain decoding}, in:
  \bibinfo{booktitle}{IEEE PRNI}, \bibinfo{address}{Stanford, CA, USA}. pp.
  \bibinfo{pages}{5 -- 8}.
\bibitem[{Onal et~al.(2015b)Onal, Ozay and Yarman~Vural}]{Onal2015_2}
\bibinfo{author}{Onal, I.}, \bibinfo{author}{Ozay, M.},
  \bibinfo{author}{Yarman~Vural, F.T.}, \bibinfo{year}{2015}b.
\newblock \bibinfo{title}{Functional mesh model with temporal measurements for
  brain decoding}, in: \bibinfo{booktitle}{EMBC}, pp.
  \bibinfo{pages}{2624--2628}.
\bibitem[{Oneata et~al.(2013)Oneata, Verbeek and Schmid}]{Oneta2013}
\bibinfo{author}{Oneata, D.}, \bibinfo{author}{Verbeek, J.},
  \bibinfo{author}{Schmid, C.}, \bibinfo{year}{2013}.
\newblock \bibinfo{title}{Action and event recognition with fisher vectors on a
  compact feature set}, in: \bibinfo{booktitle}{ICCV}, pp.
  \bibinfo{pages}{1817--1824}.
\bibitem[{Ozay et~al.(2012)Ozay, {\"{O}}ztekin, {\"{O}}ztekin and
  Yarman{-}Vural}]{mads}
\bibinfo{author}{Ozay, M.}, \bibinfo{author}{{\"{O}}ztekin, I.},
  \bibinfo{author}{{\"{O}}ztekin, U.}, \bibinfo{author}{Yarman{-}Vural, F.T.},
  \bibinfo{year}{2012}.
\newblock \bibinfo{title}{Mesh learning for classifying cognitive processes}.
\newblock \bibinfo{journal}{CoRR} \bibinfo{volume}{abs/1205.2382}.
\bibitem[{Perronnin et~al.(2010)Perronnin, S{\'a}nchez and
  Mensink}]{Perronnin2010}
\bibinfo{author}{Perronnin, F.}, \bibinfo{author}{S{\'a}nchez, J.},
  \bibinfo{author}{Mensink, T.}, \bibinfo{year}{2010}.
\newblock \bibinfo{title}{Improving the fisher kernel for large-scale image
  classification}, in: \bibinfo{booktitle}{ECCV}.
  \bibinfo{publisher}{Springer}, pp. \bibinfo{pages}{143--156}.
\bibitem[{Richiardi et~al.(2011)Richiardi, Eryilmaz, Schwartz, Vuilleumier and
  Ville}]{Richiardi2011}
\bibinfo{author}{Richiardi, J.}, \bibinfo{author}{Eryilmaz, H.},
  \bibinfo{author}{Schwartz, S.}, \bibinfo{author}{Vuilleumier, P.},
  \bibinfo{author}{Ville, D.V.D.}, \bibinfo{year}{2011}.
\newblock \bibinfo{title}{Decoding brain states from fmri connectivity graphs}.
\newblock \bibinfo{journal}{NeuroImage} \bibinfo{volume}{56},
  \bibinfo{pages}{616 -- 626}.
\bibitem[{S{\'a}nchez et~al.(2013)S{\'a}nchez, Perronnin, Mensink and
  Verbeek}]{Sanchez2013}
\bibinfo{author}{S{\'a}nchez, J.}, \bibinfo{author}{Perronnin, F.},
  \bibinfo{author}{Mensink, T.}, \bibinfo{author}{Verbeek, J.},
  \bibinfo{year}{2013}.
\newblock \bibinfo{title}{Image classification with the fisher vector: Theory
  and practice}.
\newblock \bibinfo{journal}{Int. J. Comput. Vis.} \bibinfo{volume}{105},
  \bibinfo{pages}{222--245}.
\bibitem[{S{\'a}nchez and Redolfi(2015)}]{sanchez2015}
\bibinfo{author}{S{\'a}nchez, J.}, \bibinfo{author}{Redolfi, J.},
  \bibinfo{year}{2015}.
\newblock \bibinfo{title}{Exponential family fisher vector for image
  classification}.
\newblock \bibinfo{journal}{Pattern Recognition Letters} \bibinfo{volume}{59},
  \bibinfo{pages}{26--32}.
\bibitem[{Savelonas et~al.(2016)Savelonas, Pratikakis and
  Sfikas}]{Savelonas2016}
\bibinfo{author}{Savelonas, M.A.}, \bibinfo{author}{Pratikakis, I.},
  \bibinfo{author}{Sfikas, K.}, \bibinfo{year}{2016}.
\newblock \bibinfo{title}{Fisher encoding of differential fast point feature
  histograms for partial 3d object retrieval}.
\newblock \bibinfo{journal}{Pattern Recognition} \bibinfo{volume}{55},
  \bibinfo{pages}{114--124}.
\bibitem[{Sekma et~al.(2015)Sekma, Mejdoub and Amar}]{Sekma2015}
\bibinfo{author}{Sekma, M.}, \bibinfo{author}{Mejdoub, M.},
  \bibinfo{author}{Amar, C.B.}, \bibinfo{year}{2015}.
\newblock \bibinfo{title}{Human action recognition based on multi-layer fisher
  vector encoding method}.
\newblock \bibinfo{journal}{Pattern Recognition Letters} \bibinfo{volume}{65},
  \bibinfo{pages}{37--43}.
\bibitem[{Shirer et~al.(2011)Shirer, Ryali, Rykhlevskaia, Menon and
  Greicius}]{Shirer2011}
\bibinfo{author}{Shirer, W.R.}, \bibinfo{author}{Ryali, S.},
  \bibinfo{author}{Rykhlevskaia, E.}, \bibinfo{author}{Menon, V.},
  \bibinfo{author}{Greicius, M.D.}, \bibinfo{year}{2011}.
\newblock \bibinfo{title}{Decoding subject-driven cognitive states with
  whole-brain connectivity patterns}.
\newblock \bibinfo{journal}{Cerebral Cortex} .
\bibitem[{Simonyan et~al.(2013)Simonyan, Parkhi, Vedaldi and
  Zisserman}]{Simonyan2013}
\bibinfo{author}{Simonyan, K.}, \bibinfo{author}{Parkhi, O.M.},
  \bibinfo{author}{Vedaldi, A.}, \bibinfo{author}{Zisserman, A.},
  \bibinfo{year}{2013}.
\newblock \bibinfo{title}{Fisher vector faces in the wild.}, in:
  \bibinfo{booktitle}{BMVC}, p.~\bibinfo{pages}{11}.
\bibitem[{Solmaz et~al.(2012)Solmaz, Dey, Rao and Shah}]{Solmaz2012}
\bibinfo{author}{Solmaz, B.}, \bibinfo{author}{Dey, S.}, \bibinfo{author}{Rao,
  A.R.}, \bibinfo{author}{Shah, M.}, \bibinfo{year}{2012}.
\newblock \bibinfo{title}{Adhd classification using bag of words approach on
  network features}, in: \bibinfo{booktitle}{SPIE Medical Imaging}, pp.
  \bibinfo{pages}{83144T--83144T}.
\bibitem[{Tzourio-Mazoyer et~al.(2002)Tzourio-Mazoyer, Landeau, Papathanassiou,
  Crivello, Etard, Delcroix, Mazoyer and Joliot}]{Tzourio2002}
\bibinfo{author}{Tzourio-Mazoyer, N.}, \bibinfo{author}{Landeau, B.},
  \bibinfo{author}{Papathanassiou, D.}, \bibinfo{author}{Crivello, F.},
  \bibinfo{author}{Etard, O.}, \bibinfo{author}{Delcroix, N.},
  \bibinfo{author}{Mazoyer, B.}, \bibinfo{author}{Joliot, M.},
  \bibinfo{year}{2002}.
\newblock \bibinfo{title}{Automated anatomical labeling of activations in spm
  using a macroscopic anatomical parcellation of the mni mri single-subject
  brain}.
\newblock \bibinfo{journal}{Neuroimage} \bibinfo{volume}{15},
  \bibinfo{pages}{273--289}.
\bibitem[{Wang et~al.(2013)Wang, Liu, She, Nahavandi and Kouzani}]{Wang2013}
\bibinfo{author}{Wang, J.}, \bibinfo{author}{Liu, P.}, \bibinfo{author}{She,
  M.F.}, \bibinfo{author}{Nahavandi, S.}, \bibinfo{author}{Kouzani, A.},
  \bibinfo{year}{2013}.
\newblock \bibinfo{title}{Bag-of-words representation for biomedical time
  series classification}.
\newblock \bibinfo{journal}{Biomedical Signal Processing and Control}
  \bibinfo{volume}{8}, \bibinfo{pages}{634--644}.
\bibitem[{Xia et~al.(2013)Xia, Wang and He}]{Xia2013}
\bibinfo{author}{Xia, M.}, \bibinfo{author}{Wang, J.}, \bibinfo{author}{He,
  Y.}, \bibinfo{year}{2013}.
\newblock \bibinfo{title}{Brainnet viewer: a network visualization tool for
  human brain connectomics}.
\newblock \bibinfo{journal}{PloS one} \bibinfo{volume}{8},
  \bibinfo{pages}{e68910}.
\bibitem[{Zhou et~al.(2016)Zhou, Wang, l.~liu, Ogunbona and Shen}]{Zhou2016}
\bibinfo{author}{Zhou, L.}, \bibinfo{author}{Wang, L.},
  \bibinfo{author}{l.~liu}, \bibinfo{author}{Ogunbona, P.},
  \bibinfo{author}{Shen, D.}, \bibinfo{year}{2016}.
\newblock \bibinfo{title}{Learning discriminative bayesian networks from
  high-dimensional continuous neuroimaging data}.
\newblock \bibinfo{journal}{IEEE Trans. Pattern Anal. Mach. Intell.} .

\end{thebibliography}

\end{document}